\documentclass[10pt,journal,compsoc]{IEEEtran}
\usepackage{cite}
\usepackage{amsmath,amssymb,amsfonts}
\usepackage{algorithmic}
\usepackage{graphicx}
\usepackage{textcomp}
\usepackage{comment}
\usepackage{threeparttable}
\usepackage{hyperref}
\def\BibTeX{{\rm B\kern-.05em{\sc i\kern-.025em b}\kern-.08em
    T\kern-.1667em\lower.7ex\hbox{E}\kern-.125emX}}
\begin{document}
\title{Skeleton-Based Action Segmentation with Multi-Stage Spatial-Temporal Graph Convolutional Neural Networks}
\author{Benjamin Filtjens, Bart Vanrumste, Peter Slaets
\thanks{Submitted on x August 2022}
\thanks{Benjamin Filtjens and Peter Slaets are with the Department of Mechanical Engineering,
KU Leuven, 3001 Leuven, Belgium. (email: benjamin.filtjens@kuleuven.be and peter.slaets@kuleuven.be).}
\thanks{Benjamin Filtjens and Bart Vanrumste are with the Department of Electrical Engineering (ESAT),
KU Leuven, 3001 Leuven, Belgium. (email: benjamin.filtjens@kuleuven.be and bart.vanrumste@kuleuven.be).}}

\maketitle

\begin{abstract}
The ability to identify and temporally segment fine-grained actions in motion capture sequences is crucial for applications in human movement analysis. Motion capture is typically performed with optical or inertial measurement systems, which encode human movement as a time series of human joint locations and orientations or their higher-order representations. State-of-the-art action segmentation approaches use multiple stages of temporal convolutions. The main idea is to generate an initial prediction with several layers of temporal convolutions and refine these predictions over multiple stages, also with temporal convolutions. Although these approaches capture long-term temporal patterns, the initial predictions do not adequately consider the spatial hierarchy among the human joints. To address this limitation, we recently introduced multi-stage spatial-temporal graph convolutional neural networks (MS-GCN). Our framework replaces the initial stage of temporal convolutions with spatial graph convolutions and dilated temporal convolutions, which better exploit the spatial configuration of the joints and their long-term temporal dynamics. Our framework was compared to four strong baselines on five tasks. Experimental results demonstrate that our framework is a strong baseline for skeleton-based action segmentation. 
\end{abstract}


\section{Introduction}
\label{sec:introduction}
\IEEEPARstart{T}{he} automatic identification and localisation of events and actions in long untrimmed motion capture (MoCap) sequences are crucial for various use-cases in human movement analysis. Typically, MoCap is performed with optical or inertial measurement systems, which encode human movement as a time series of human joint locations and orientations or their higher-order representations \cite{Al-Amri2018-yd, Mahmood2019-au}. The high-dimensional time series registers the articulated motion as a high degree of freedom human skeleton. Therefore, MoCap sequences can be generically regarded as skeleton-like inputs. Given an untrimmed skeleton sequence, we aim to segment every event and action in time. In the literature, this task falls under the domain of skeleton-based action segmentation. \\
Related to this task is the task of skeleton-based action recognition. Unlike action segmentation, action recognition aims to classify actions from short and well-segmented video clips. This domain has made tremendous strides due to the availability of low-cost MoCap approaches. These approaches are driven by pose estimation algorithms, which are a form of marker-less optical MoCap that encode human movement as a time series of human joint locations with a single camera \cite{Shotton2013-zx, Cao2018-yi}. Human actions can then be recognized by appropriately modelling the high dimensional time series. Earlier methods ignored the spatial hierarchy among the joints and modelled human actions by applying high-level temporal models \cite{Fernando2015-hb}. Later methods explicitly modelled the natural connection between joints \cite{Shahroudy2016-sz}. These methods showed encouraging improvement, which suggests the significance of modelling the spatial hierarchy among the joints. The state-of-the-art approaches are based on the spatial-temporal graph convolutional neural network (ST-GCN) \cite{Yan2018-jp}. These approaches model the skeleton sequences as a spatial-temporal graph. The idea is to construct a graph in which each node corresponds to a human body joint and the edges correspond to the spatial connectivity among the joints and the temporal connectivity of the same joint across time. The spatial-temporal graph can then be modelled by graph neural networks, which generalize convolutional neural networks to graphs of arbitrary structures \cite{Defferrard2016-cb, Kipf2017}. However, skeleton-based action segmentation is more challenging than recognition, due to the need for simultaneous recognition and localization. Despite its broad potential in human movement analysis, a proper framework for this task has not yet been established. \\
Within the generic domain of action segmentation, i.e. approaches that are not specifically designed for skeleton data, earlier methods mainly utilized a sliding-window scheme \cite{Singh2016-iy, Sun2015-ny}. However, the optimal window size is often a trade-off between model expressivity, i.e. the models' ability to capture long-term temporal context, and the sensitivity of the model to take into account short actions \cite{Yao2018-iz}. Recent methods, such as temporal convolutional neural networks (TCN) \cite{Lea2017}, can operate on untrimmed sequences and classify each time sample, termed action segmentation, for simultaneous action recognition and localisation. TCNs perform dilated temporal convolutions to capture long-term temporal context \cite{Yu2015-qu}. In action segmentation the predictions tend to vary at a high temporal frequency, often resulting in over-segmentation errors. To address this problem, the state of the art approach, termed multi-stage temporal convolutional neural networks (MS-TCN), includes refinement stages \cite{Farha2019-yw}. The idea is to employ a temporal model to generate an initial prediction and refine these predictions over multiple stages. However, these generic action segmentation approaches do not consider the spatial hierarchy among the skeleton joints. \\
We recently introduced an architecture for clinical freezing of gait (FOG) assessment in Parkinson's disease based on optical marker-based motion capture data, termed multi-stage spatial-temporal graph convolutional neural network (MS-GCN) \cite{Filtjens2021-hu}. Our architecture amalgamates the best practices in convolutional neural network design to address the task of skeleton-based action segmentation. First, we extended ST-GCN for action segmentation by including dilation on the temporal graph to increase the temporal receptive field \cite{Yu2015-qu}. Next, we modified MS-TCN by decoupling the prediction generation stage from the refinement stages, allowing us to address the different goals of these stages. Specifically, we replaced the TCN-based temporal layers that generate an initial prediction by the modified ST-GCN layers to appropriately model the spatial hierarchy among the joints. We hypothesize that MS-GCN is a strong baseline for skeleton-based action segmentation tasks other than FOG assessment and for other MoCap representations than optical marker-based MoCap. To this end, the contribution of the present manuscript is four-fold: (1) We propose MS-GCN as a generic baseline for skeleton-based action segmentation. (2) We introduce five relevant use-cases from four public datasets and one proprietary dataset. The use-cases include three different forms of motion capture,  marker-based and marker-less optical MoCap, and inertial-based MoCap. (3) We show that the proposed architecture exceeds the performance of four strong deep learning baseline methods. (4) We publicly release our code and trained models at: \url{https://github.com/BenjaminFiltjens/MS-GCN}.

\section{Skeleton-based action segmentation}
This section first formalizes the problem of skeleton-based action segmentation. Next, we introduce the three distinguishing characteristics of the MS-GCN architecture, which are: (1) dilated temporal convolutions to learn long-term temporal patterns \cite{Lea2017}, (2) spatial graph convolutions to learn spatial patterns \cite{Yan2018-jp}, (3) multiple stages of refinement to reduce the number of segmentation errors \cite{Farha2019-yw}. These characteristics are further discussed within this section.

\subsection{Problem statement}
A MoCap sequence can be generically represented as: $f \in \mathbb{R}^{T\times N\times C}$, where $T$ are the number of samples, $N$ are the number of nodes, and $C$ are the number of feature channels per node. Note that the number of samples $T$ may vary for each input sequence. Given a MoCap sequence, we aim to infer the class label for each sample $\hat{Y} = \hat{y}_{0}, \dots, \hat{y}_{T}$. The inferred class labels are represented as: $\hat{Y} \in \mathbb{R}^{T\times L}$, where $\hat{y}_{t,l}$ is the probability of class $l$ at sample $t$. 

\subsection{Dilated temporal convolution}
\begin{figure}
\centering
\includegraphics[width=2.5in]{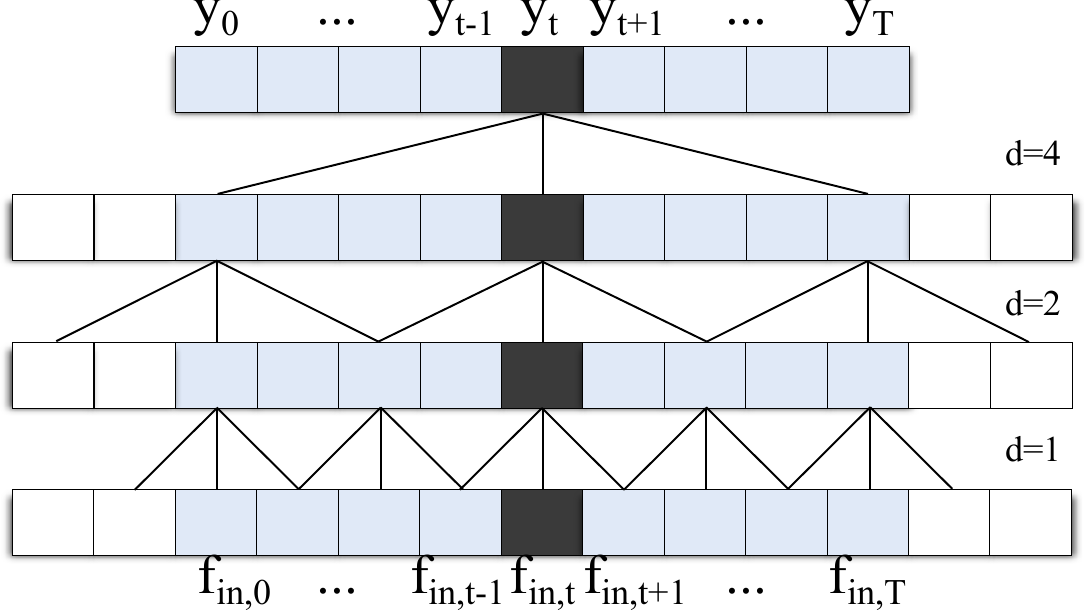}
\caption{Visual overview of a (dilated) temporal convolutional neural network (TCN). The visualized network is implemented in acausal mode, since the filters take into account future observations $f_{in,t+1}, \dots, f_{in,T}$. The first layer has a dilation rate of 1, reducing this layer to a regular convolution. By increasing the dilation rate $d$ throughout the network, the deeper layers can represent a wider range of inputs, thereby expanding the temporal receptive field of the network.}
\label{fig_dil}
\end{figure}

Convolutional neural networks (CNN) are ideal for processing data with a grid-like topology such as time-series (1D CNN) and images (2D CNN) \cite{Goodfellow2016-tq}. A CNN learns an expressive representation through altering convolutional and pooling layers \cite{Lecun1998-an}. The pooling layers downsample the temporal representation, allowing the model to capture long-range dependencies at the cost of losing fine-grained information. Recent temporal convolutional neural networks (TCN) omit pooling and instead use dilated convolutions \cite{Yu2015-qu} to capture long-range dependencies while keeping the temporal representation intact \cite{Farha2019-yw}. For an input feature map $f_{in}$ and a filter $p$, the dilated convolution on sample $t$ of the feature map is defined as \cite{Bai2018-uk}:
\begin{equation}
(f_{in} *_d p)(t) = \sum_{i=0}^{k-1}p(i) \cdot f_{in_{t-d \cdot i}},
\end{equation}\label{equationtcn}
where $*_d$ is the dilated convolution operator with dilation rate $d$, $k$ is the size of the filter (kernel), and $t-d \cdot i$ is used to indicate that the filter in Equation 1 is applied in causal mode, i.e. direction of the past. The filter can be implemented in acausal mode, i.e. take into account future observations, by zero-padding symmetrically. By increasing the dilation rate $d$ throughout the network, the deeper layers can represent a wider range of inputs, thereby expanding the temporal receptive field of the network. A visual overview of a dilated TCN is provided in Figure \ref{fig_dil}.

\subsection{Graph convolution}
\begin{figure}
\centering
\includegraphics[width=3.2in]{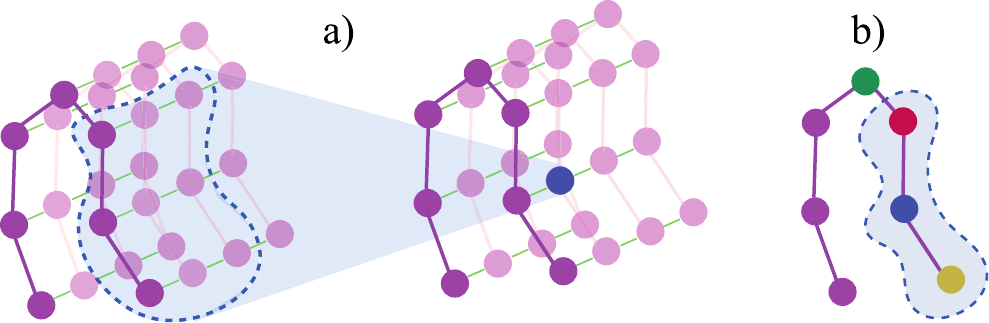}
\caption{Spatial-temporal graph convolutional neural network (ST-GCN). Visual overview of a spatial-temporal graph (a) and spatial partitioning strategy (b). The spatial partitioning strategy has three subsets based on a nodes distance with respect to a self-selected root node (green). The three subsets are the node itself (blue), the node closest to the root node (red), and the node furthest from the root node (yellow).}
\label{fig_gcn}
\end{figure}

Graph convolutional neural networks (GCNs) generalize CNNs to non-euclidian structured data \cite{Kipf2017}. Yan et al. extended GCNs to exploit the inherent spatial relationship among the joints of a skeleton \cite{Yan2018-jp}. Their approach termed spatial-temporal graph convolutional networks (ST-GCN) learns a representation on a graph $G = (V, E, A)$ which takes as input:
\begin{itemize}
    \item A set of nodes $V = \{v_{ti}|t = 1,...,T, i = 1,...,N\}$ for a skeleton sequence of $N$ joints and $T$ samples.
    \item Two sets of edges $E_S = \{v_{ti}v_{tj}|(i,j) \in H\}$ and $E_F = \{v_{ti}v_{(t+1)i}\}$, where $H$ is the set of connected joints. $E_S$ refers to the intra-skeleton edges at each frame (spatial dimension), and $E_F$ refers to the inter-frame connection of the same joints over all of the frames (temporal dimension).
    \item A description of the graph structure in the form of an adjacency matrix $A$.
\end{itemize}
For instance, Figure \ref{fig_gcn}(a) visualizes the spatial-temporal graph. The joints represent the nodes of the graph (purple nodes), their natural connections are the spatial edges (purple lines), and the connection between adjacent frames are the temporal edges (green lines). \\
In the spatial dimension, the graph convolution operation on node $v_{ti}$ is defined as \cite{Yan2018-jp}:
\begin{equation}
    f_{gcn}(v_{ti}) = \sum_{v_{tj} \in B(v_{ti})} \frac{1}{Z_{ti}(v_{tj})}f_{in}(v_{tj})\cdot w(l_{ti}(v_{tj})),
\end{equation}
where $f_{in}$ and $f_{gcn}$ denote the input feature map and output feature map, respectively. The term $B(v_{ti})$ denotes the sampling area of node $v_{ti}$, with the nodes within the sampling area denoted as $v_{tj}$. A mapping function $l_{ti}$ is defined to map each node with a unique weight vector $w$. Figure \ref{fig_gcn}(b) visualizes this strategy for a single frame $t$, where the kernel size is set as 3 and the sampling area $B$ is partitioned into 3 subsets based on a nodes distance with respect to a self-selected root node (green). The three subsets in this partitioning strategy are the node itself (blue), the node closer to the root node (red), and the node further from the root node (yellow). The normalizing term $Z_{ij}$ is added to balance the contributions of different subsets to the output. 

\subsection{Refinement stages}
As predictions are made at high temporal frequencies, over-segmentation errors, i.e. an action is segmented into multiple shorter actions, often occur. A common strategy to alleviate this problem in pixel-wise labelling of images is to generate an initial prediction, then refine this initial prediction using the interactions between neighbouring pixels \cite{Goodfellow2016-tq}. Farha and Gall extend this to action segmentation in time series data \cite{Farha2019-yw}. The idea is to stack several predictors that each operates directly on the output of the previous one to incrementally refine the predictions.

\section{Deep learning models}\label{impl}
\begin{figure*}[!t]
\centering
\includegraphics[width=7in]{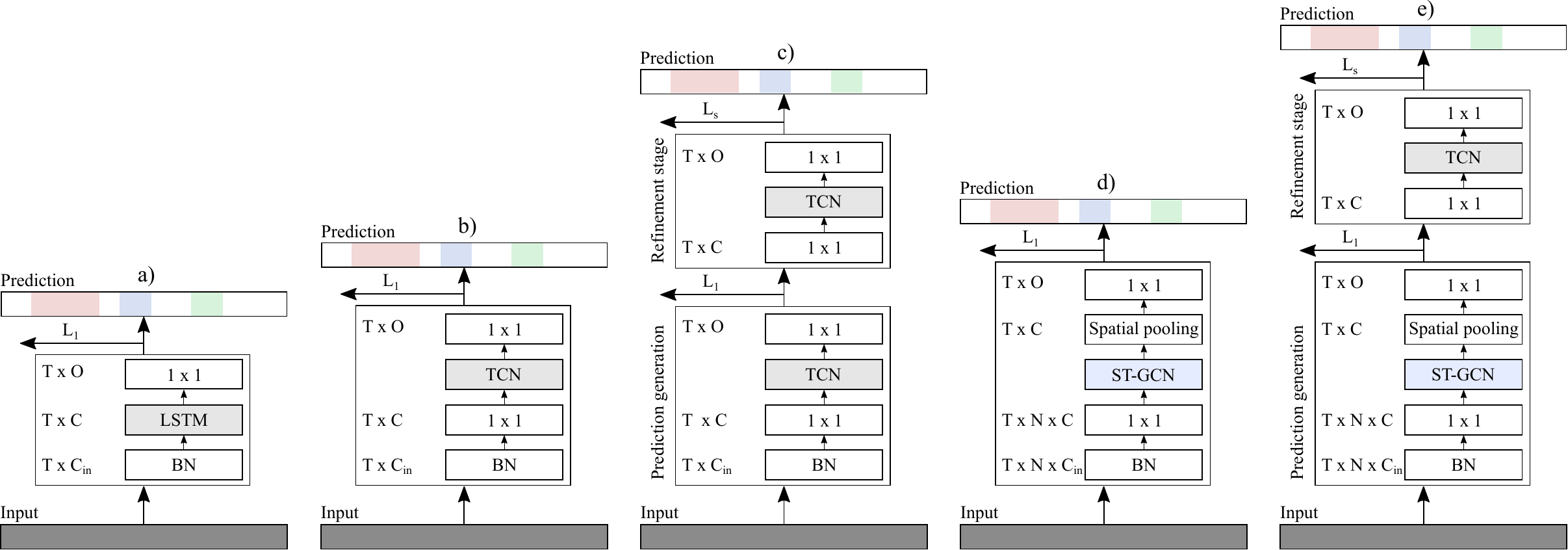}
\caption{Overview of the MS-GCN and the four baseline deep learning models. The models take as input a MoCap sequence and generate as output a sequence of actions. The five deep learning models are: (a) a long short-term memory network (LSTM), (b) a temporal convolutional neural network (TCN), (c) a multi-stage temporal convolutional neural network (MS-TCN), (d) a spatial-temporal graph convolutional neural network (ST-GCN), and (e) a multi-stage spatial-temporal graph convolutional neural network (MS-GCN). The terms BN and $L_s$ denote the batch normalization layer and the loss of stage $s$, respectively.}
\label{fig_models}
\end{figure*}

The previous section introduced the three building blocks that characterizes the MS-GCN architecture. As the MS-GCN architecture combines the best practices from TCN, ST-GCN, and MS-TCN, we include these as a baseline. We additionally include a bidirectional long short term memory-based network (LSTM) \cite{Graves2005-gv}, as it is often considered an important baseline in action segmentation of MoCap data \cite{Filtjens2020-hl, Matsushita2021-kx, Cheema2018-sh}. \\
The implementation details of the employed models are visualized in Figure \ref{fig_models}. The first layer of all models is a batch normalization (BN) layer that normalizes the inputs and accelerates training \cite{Ioffe2015-ta}. After normalization, the input is reshaped into the accepted formats of the specified models. For the graph-based models (i.e., ST-GCN and MS-GCN), the data is shaped into $T \times N \times C_{in}$, where $N$ represents the number of nodes, $C_{in}$ the number of input channels, and T the number of samples. For the temporal models (i.e., LSTM, TCN, and MS-TCN), the data is shaped into $T \times C_{in}N$. For these models, all input node locations are thus concatenated to form the input features at each sample $t$. 

\subsection{LSTM}
The first layer of our recurrent model is an LSTM layer, which computes the following function:
\[i_{t} = \sigma(f_{in_t}W_{ii} + b_{ii} + h_{t-1}W_{hi} + b_{hi}),\]
\[j_{t} = \sigma(f_{in_t}W_{if} + b_{if} + h_{t-1}W_{hf} + b_{hf}),\]
\[\tilde{c}_{t} = tanh(f_{in_t}W_{ic} + b_{ic} + h_{t-1}W_{hc} +  b_{hc}),\]
\[o_{t} = \sigma(f_{in_t}W_{io} +  b_{io} + h_{t-1}W_{ho} +  b_{ho}),\]
\[c_{t} = j_{t} \odot c_{t-1} + i_{t} \odot \tilde{c}_{t}),\]
\[h_{t} = tanh(c_{t}) \odot o_{t},\]
where $h_t$ is the hidden state at sample $t$, $c_t$ is the cell state at sample $t$, $f_{in_t}$ is the input feature map at sample t, $h_{t-1}$ is the hidden state of the layer at sample $t-1$. The terms $i_t$, $j_t$, and $o_t$ are the input, forget, and output gates, respectively. The terms $\sigma$, $tanh$, and $\odot$ are the sigmoid function, hyperbolic tangent function, and Hadamard product, respectively. The weight matrices are represented by $W$, with subscripts representing from-to relationships. The LSTM layer above is causal, as the hidden state $h_t$ depends only on $x_{0}, \dots, x_{t}$. The LSTM can be implemented in acausal mode, i.e. take into account future observations $x_{t+1}, \dots, x_{T}$, by training it in the positive and negative time direction (bidirectional) \cite{Schuster1997-nd, Graves2005-gv}. The hidden representation of the past and future are then combined through simple concatenation \cite{Graves2005-gv}. A visual overview of the (bidirectional) LSTM network is provided in Figure \ref{fig_models}(a). 

\subsection{TCN}\label{ssec:tcn}
\begin{figure}
\centering
\includegraphics[width=2in]{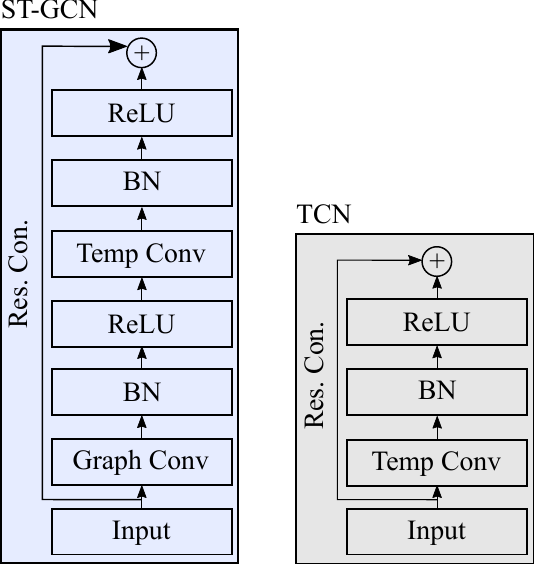}
\caption{Visual overview of a temporal convolutional (TCN) and spatial-temporal graph convolutional (ST-GCN) block \cite{Yan2018-jp}. ST-GCN generates a spatial-temporal feature map by applying a spatial graph convolution (see Figure \ref{fig_gcn}(b)) and a temporal convolution (see Figure \ref{fig_dil}), both of which are followed by batch normalization (BN) and a ReLU non-linearity. Moreover, a residual connection is added to each block.}
\label{fig_st-gcn}
\end{figure}
The first layer of the TCN-based model is a $1 \times 1$ convolutional layer that adjusts the input dimension $C_{in}$ to the number of filters $C$ in the network, formalized as:
\begin{equation}
f_{adj} = W_1 * f_{in}+b,
\end{equation}
where $f_{adj} \in \mathbb{R}^{T\times C}$ is the adjusted feature map, $f_{in} \in \mathbb{R}^{T\times C_{in}}$ the input MoCap sequence, $*$ the convolution operator, $b \in \mathbb{R}^{C}$ the bias term, and $W_1 \in \mathbb{R}^{1\times C_{in}\times C}$ the weights of the $1\times 1$ convolution filter with $C_{in}$ input feature channels and $C$ equal to the number of feature channels in the network. \\
The adjusted input is passed through several TCN blocks (visualized in \ref{fig_st-gcn}). Each TCN block applies a dilated temporal convolution \cite{Yu2015-qu}, BN, ReLU non-linear activation, and a residual connection between the activation map and the input. Formally, this process is defined as:
\begin{equation}
f_{out} = \delta(BN(W *_d f_{adj}+b)) + f_{adj},
\end{equation}
where $f_{out} \in \mathbb{R}^{T\times C}$ is the output feature map, $*_d$ the dilated convolution operator, $b \in \mathbb{R}^{C}$ the bias term, $W \in \mathbb{R}^{k\times C\times C}$ the weights of the dilated convolution filter with kernel size $k$, and $\delta$ the ReLU function. A visual overview of the TCN-based network is provided in Figure \ref{fig_models}(b). 

\subsection{ST-GCN}\label{ssec:st-gcn}
The first layer of the ST-GCN-based model is a $1 \times 1$ convolutional layer that adjusts the input dimension $C_{in}$ to the number of filters $C$ in the network, formalized as:
\begin{equation}
f_{adj} = W_1 * f_{in}+b,
\end{equation}
where $f_{adj} \in \mathbb{R}^{T\times N\times C}$ is the adjusted feature map, $f_{in} \in \mathbb{R}^{T\times N\times C_{in}}$ the input MoCap sequence, $*$ the convolution operator, $b \in \mathbb{R}^{C}$ the bias term, $W_1 \in \mathbb{R}^{1\times 1\times C_{in}\times C}$ the weights of the $1\times1$ convolution filter with $C_{in}$ input feature channels and $C$ equal to the number of feature channels in the network. \\
The adjusted input is passed through several ST-GCN blocks (visualized in \ref{fig_st-gcn}) \cite{Yan2018-jp}. Each ST-GCN first applies a graph convolution, transforming Equation 2 into:
\begin{equation}
f_{gcn} = \sum_{p} A_p f_{adj}W_p M_p,
\end{equation}
where $f_{adj} \in \mathbb{R}^{T \times N \times C}$ is the adjusted input feature map, $f_{gcn} \in \mathbb{R}^{T \times N \times C}$ the output feature map of the spatial graph convolution, and $W_p$ the $1 \times 1 \times C \times C$ weight matrix. The matrix ${A_p} \in \{0,1\}^{N\times N}$ is the adjacency matrix, which represents the spatial connection between the joints. The adjacency matrix $A_p = D_p^{-\frac{1}{2}}\: A_p\: D_p^{-\frac{1}{2}}$, where $D_p$ is the diagonal node degree matrix. Multiplying $D_p^{-\frac{1}{2}}\: A_p\: D_p^{-\frac{1}{2}}$ corresponds to symmetrically normalizing $A$, which prevents changing the scale of the features based on the number of connections \cite{Kipf2017}. The graph is partitioned into three subsets $p$ based on the spatial partitioning strategy, as was visualized in Figure \ref{fig_gcn}(b) \cite{Yan2018-jp}. There are thus three different weight vectors $W_p$ that allow modelling of relative properties between the nodes. The matrix $M_p$ is a learnable ${N\times N}$ attention mask that indicates the importance of each node and its spatial partitions. \\
Next, after passing through a BN layer and ReLu non-linearity, the ST-GCN block performs a dilated temporal convolution. The dilated temporal convolution is, in turn, passed through a BN layer and ReLU non-linearity, and lastly, a residual connection is added between the activation map and the input. This process is formalized as:
\begin{equation}
f_{out} = \delta(BN(W *_d f_{gcn}+b)) + f_{adj},
\end{equation}
where $f_{out} \in \mathbb{R}^{T\times N\times C}$ is the output feature map, $*_d$ the dilated convolution operator, $b \in \mathbb{R}^{C}$ the bias term, $W \in \mathbb{R}^{k \times 1\times C\times C}$ the weights of the dilated convolution filter with kernel size $k$. The output feature map is passed through a spatial pooling layer that aggregates the spatial features among the $N$ joints. A visual overview of the ST-GCN-based network is provided in Figure \ref{fig_models}(d). 

\subsection{Single-stage models: sample-based prediction}
The three aforementioned single-stage models map an input skeleton sequence $f_{in}$ to a hidden representation $f_{out} \in \mathbb{R}^{T \times C}$, with $C$ determined by the number of convolutional filters (ST-GCN and TCN) or the number of hidden units (LSTM), and length $T$ the same as the input sequence. The hidden representation of each model is passed through a $1 \times 1$ convolution and a softmax activation function to get the probabilities for the $L$ output classes for each sample in-time, formalized as:
\begin{equation}
\hat{Y} = \zeta(W_1 * f_{out} + b),
\end{equation}
where $\hat{Y} \in \mathbb{R}^{T\times L}$ are the class probabilities at each sample $t$, $f_{out}$ the hidden output representation of the single stage models, $*$ the convolution operator, $b \in \mathbb{R}^{L}$ the bias term, $\zeta$ the softmax function, $W_1 \in \mathbb{R}^{1\times C \times L}$ the weights of the $1\times 1$ convolution filter with $C$ input channels and $L$ output classes.

\subsection{Multi-stage models: prediction refinement}
The initial predictions predictions $\hat{Y} \in \mathbb{R}^{T\times L}$ are passed through several refinement stages. Each refinement stage contains several TCN blocks, and each stage operates directly on the softmax activations of the previous stage. Formally, this process is defined as:
\begin{equation}
\hat{Y}^{s} = \Gamma(\hat{Y}^{s-1}),
\end{equation}
where $\hat{Y}^{s} \in \mathbb{R}^{T\times L}$ is the output at stage $s$, $\hat{Y}^{s-1}$ the output of the previous stage, and $\Gamma$ the single-stage TCN, as explained in section \ref{ssec:tcn}. \\
For the MS-TCN architecture, the initial predictions are generated by the single-stage TCN discussed in Chapter \ref{ssec:tcn}. For the MS-GCN architecture, the initial predictions are generated by the single-stage ST-GCN discussed in Chapter \ref{ssec:st-gcn}. A visual overview of the MS-TCN and MS-GCN network are provided in Figure \ref{fig_models}(c) and (e), respectively. 

\subsection{Implementation details}
\subsubsection{Loss function}
The models were trained by minimizing a combined cross-entropy (CE) and mean squared error (MSE) loss. The CE loss was defined as:
\begin{align}
    \mathcal{L}=\sum_{s=1}^S \mathcal{L}_{s,cls}, \\
    \mathcal{L}_{cls}=\frac{1}{T}\sum_t -y_{t,l} log(\hat{y}_{t,l}),
\end{align}
where $\mathcal{L}$ is the total loss over all $S$ stages and $\mathcal{L}_{cls}$ the CE loss with $y_{t,l}$ and  $\hat{y}_{t,l}$ the ground truth label and predicted probability for class $l$ at sample $t$, respectively. The combined CE and MSE loss was defined as \cite{Farha2019-yw}:
\begin{align}
    \mathcal{L}=\sum_{s=1}^S \mathcal{L}_{s,cls}+\lambda \mathcal{L}_{s,T-MSE},
\end{align}
where $\mathcal{L}_{T-MSE}$ is the MSE loss and $\lambda$ is a hyperparameter that determines its contribution. The combined loss was proposed by Farha and Gall to avoid over-segmentation errors \cite{Farha2019-yw}, which occur when predictions vary at an unrealistically high sample frequency. The MSE term negates this effect by calculating the truncated mean squared error over the sample-wise log probabilities. The MSE loss function is defined as:
\begin{align}
    \mathcal{L}_{T-MSE}=\frac{1}{TL}\sum_{t,l}\widetilde{\Delta}_{t,l}^2,\\
    \widetilde{\Delta}_{t,l}=
        \begin{cases}
        \Delta_{t,l} & :\Delta_{t,l}\leq\tau\\
        \tau & : otherwise
        \end{cases},\\
        \Delta_{t,l}=|log(\hat{y}_{t,l})-log(\hat{y}_{t-1,l})|,
\end{align} 
where $T$ is the sequence length, $L$ is the number of classes, and $\hat{y}_{t,l}$ is the probability of class $l$ at sample $t$. The hyperparameter $\tau$ defines the threshold to truncate the smoothing loss. 
\subsubsection{Model hyperparameters}
To avoid model selection bias for the convolutional models, (i.e., TCN, ST-GCN, MS-TCN, and MS-GCN), the same model hyperparameters were chosen as MS-TCN \cite{Farha2019-yw}. More specifically, each layer had 64 filters with a temporal kernel size of 3. All multi-stage models had 1 prediction generation stage and 3 refinement stages, and each stage had 10 layers. The convolutions were acausal, i.e. they could take into account both past and future input features. The dilation factor of the temporal convolutions doubled at each layer, i.e. 1, 2, 4, ..., 512. \\
For the recurrent model, we followed a configuration that is conventional in MoCap-based action segmentation. For instance, prior work in gait cycle and FOG subtask segmentation used recurrent models of 1-3 LSTM layers of 32 - 128 cells each \cite{Kidzinski2019-ou, Matsushita2021-kx}. For our recurrent model, we used two forward LSTM layers and two backward LSTM layers, each with 64 cells. 
\subsubsection{Optimizer hyperparameters}
The optimizer and loss hyperparameters were also selected according to MS-TCN \cite{Farha2019-yw}. For the loss, we set $\tau$ = 4 and $\lambda$ = 0.15. MS-TCN experiments show that further increasing the value of $\lambda$ and $\tau$ worsens the capability of the model in detecting the boundaries between action segments \cite{Farha2019-yw}. For the optimizer, we used Adam \cite{Kingma2014-va} with a learning rate of 0.0005. All models were trained for 100 epochs with a batch size of 4. 
\subsubsection{Ablative experiments}
For MS-GCN, we perform causal versus acausal and regular temporal convolutions versus dilated temporal convolutions ablative experiments. Causal experiments mean that the prediction at sample $t$ depends only $f_{in,0}, \dots, f_{in,t}$, which is important for real-time applications (e.g., in robotics) \cite{Lea2017}. In acausal mode the model can take into account future observations $f_{in,t+1}, \dots, f_{in,T}$, which is sufficient for post-hoc movement analysis applications. For the regular temporal convolution experiment, we set the dilation rate to 1 in each layer.

\section{Evaluation}
We present five datasets for skeleton-based action segmentation. Three of the five datasets are for action segmentation, with each featuring a different skeleton-based representation, i.e. inertial-based (HuGaDB), markerless optical MoCap (PKU-MMDv2), and marker-based optical MoCap (LARa). Two of the five datasets involve typical segmentation tasks commonly used in clinical gait analysis. For these two tasks, additional context regarding the relevance is provided.

\subsection{Peking University - Continuous Multi-Modal Human Action Understanding (PKU-MMD v2)}
PKU-MMD is a benchmark dataset for continuous 3D human action understanding \cite{Liu2017-jx}. In this study, we use the smaller phase 2 partition of the dataset. This dataset contains 1009 short video sequences in 52 action categories, performed by 13 subjects in three camera views. MoCap was performed with a Kinect v2 optical marker-less motion capture system at 30 Hz. The Kinect system records the 3-axis locations of 25 major body joints.

\subsection{Human Gait Database (HuGaDB)}
HuGaDB is an action segmentation dataset where a total of 18 subjects carried out typical lower limb activities, e.g. walking, running, and cycling \cite{Chereshnev2017HuGaDBHG}.
MoCap was performed with 6 inertial measurement units (IMUs) at a sampling frequency of 60 Hz. The IMUs were placed on the right and left thighs, shins and feet. This dataset contains 364 IMU trials in 12 action categories. 

\subsection{Logistic Activity Recognition Challenge (LARa)}
LARa is a recently released dataset of subjects carrying out typical warehousing activities \cite{Niemann2020-ut}. Fourteen subjects carried out a total of eight actions. MoCap was performed by an optical MoCap system that recorded the motion of 39 reflective markers at a sampling frequency of 200 Hz. The optical MoCap system records the 3-axis limb position and 3-axis orientation of 19 limbs. All subjects participated in a total of 30 recordings of 2 minutes each. The actions were performed under three different warehousing scenarios that each aimed to mimic real-world warehousing activities. In scenario 1, subjects 1 to 6 performed 30 recordings, and subjects 7 to 14 performed 2 recordings. Subjects 7 to 14 additionally performed 14 recordings in scenarios 2 and 3. The dataset contains 377 MoCap trials in 8 action categories. The authors proposed to tackle the automated skeleton-based segmentation task with a TCN-based model that classified temporal segments extracted by a sliding window. 

\subsection{Gait phase and freezing of gait segmentation (FOG-GAIT)}
Freezing of gait (FOG) and temporal gait disturbances in people with Parkinson's disease (PwPD) are commonly assessed during complex experimental protocols that involve turning with or without a cognitive dual-task \cite{Spildooren2010-pj, Nieuwboer2001-cr}, which serve as triggers to elicit FOG \cite{Schaafsma2003-pz}. The current assessment implies that the gait cycle phases, i.e. double support 1, single support, double support 2, and swing, and the FOG episodes are annotated manually based on the 3D marker trajectories of a motion capture system, and standard camera footage \cite{Nieuwboer2001-cr, Gilat2019-xm}. These time-consuming tasks motivate the search for algorithms to automatically delineate the gait cycle phases and FOG episodes. State-of-the-art deep learning models tackle the gait segmentation task with TCN or LSTM-based models \cite{Filtjens2020-hl, Kidzinski2019-ou}. \\
A proprietary MoCap dataset of seven PwPD and FOG that froze during the protocol was used \cite{Spildooren2010-pj}. The subjects were instructed to complete a standardized protocol consisting of straight-ahead walking, 180 degree turning, and 360 degree turning. The experiments were offered randomly and performed with our without a cognitive dual-task \cite{Bowen2001-uc}. Two optical markers were placed at a .5m distance from each other on the floor to standardize the turning radius. The data acquisition was further standardized by defining a zone of one meter before and after the turn in which MoCap data was stored. The FOG events and gait cycle phases were visually annotated by an experienced clinical operator. MoCap was performed at a sampling frequency of 100 Hz with a ten camera Vicon motion capture system. Optical markers were placed according to the plugin-gait configuration \cite{Davis1991-tr}. This dataset contains 127 MoCap trials in 5 action categories.

\subsection{Timed Up-and-Go (TUG) sub-task segmentation}
The timed up-and-go (TUG) is a commonly used test in clinical practice to evaluate a subjects' functional mobility \cite{Podsiadlo1991-gy}. During the TUG, subjects carry out several sub-activities that are common in daily life, i.e. sitting, standing up, walking, turning around, walking back, and sitting back down. In clinical practice, the timing of the sub-activities is commonly assessed under clinical supervision. Therefore, there is increased interest in automatic TUG analysis and sub-activity segmentation techniques. State-of-the-art deep learning models tackle this task with LSTM-based models \cite{Li2018-qw}. \\
We used a public dataset that aims to recruit a total of 500 healthy participants (aged 21-80) of Asian ethnicity \cite{Liang2020-hq}. At the time of this study, the data of only 10 participants were available. Each participant carried out the TUG 3 times, resulting in a total of 30 recordings. Motion capture was performed with a Qualisys optical motion capture system that recorded the motion of reflective markers at a sampling rate of 200 Hz. The markers were placed according to the modified Calibrated Anatomical System Technique (CAST) \cite{Cappozzo1995-ra}. The 6 TUG sub-activities were visually annotated by an experienced clinical operator.

\subsection{Graph representations}
\begin{figure}
\centering
\includegraphics[width=3in]{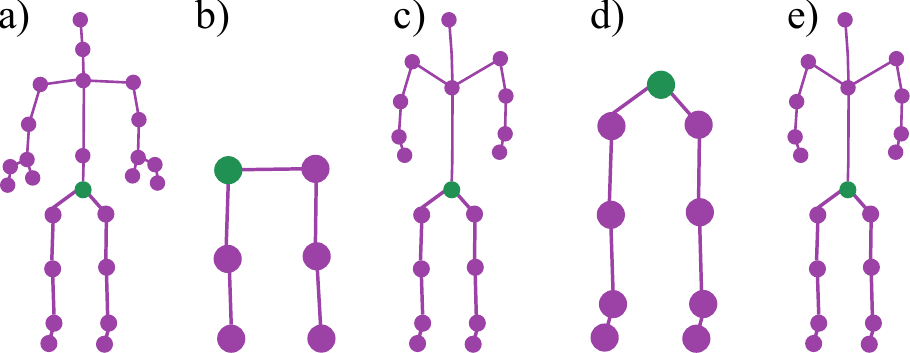}
\caption{Different graph representations for (a) PKU-MMD v2, (b) HuGaDB, (c) LARa, (d) FOG-GAIT, (e) TUG. The purple dots represent the nodes of the graph, the purple lines the spatial edges (excluding self-connections), and the green dot the root node.}
\label{fig_graphs}
\end{figure}

For each dataset, the skeleton graphs are visualized in Figure \ref{fig_graphs}. For HuGaDB we used the 3-axis accelerometer and 3-axis gyroscope data, for LARa the 3-axis limb position and 3-axis orientation, and for PKU-MMD v2, TUG, and FOG-GAIT we computed the 3-axis displacement and 3-axis relative coordinates (with respect to the root node as visualized in Figure \ref{fig_graphs}) from the 3D joint positions.

\subsection{Metrics}
\begin{table}[!h]
\caption{Dataset characteristics.}
\centering
\begin{tabular}{llllll}
Dataset       & Partitions & SR & \#N & \#Trials & \#L \\ \hline
PKU-MMD & 3/10       & 30 Hz       & 25 & 234/775 & 52 \\
HuGaDB     & 4/18         & 60 Hz       & 6 & 69/307 & 12   \\
LARa       & 4/14       & 50 Hz       & 19 & 113/264 & 8 \\
FOG-GAIT   & LOSO         & 50 Hz       & 9  & 127/127 & 5   \\
TUG        & LOSO         & 50 Hz       & 19   & 30/30 & 6     
\end{tabular}
\label{table_data}
\begin{tablenotes}
\item
Overview of the partitioning, sampling rates (SR), number of nodes (\#N), number of trials (test/train), and number of classes (\#L) across datasets. For the fixed test/train partition of HuGaDB, we selected the first four subjects as test subjects. For the fixed test/train partition of PKU-MMD v2, the 3 test subjects were provided by the authors of the dataset. For LARa, S1-6 and S7-14 perform different experiments. To cover both experiments, we take subjects 5-8 for our test partition. 
\end{tablenotes}
\end{table}

We follow convention by quantitatively evaluating the predictions with respect to the ground truth annotations by means of a sample-wise and a segment-wise evaluation metric \cite{Lea2017, Farha2019-yw}. For the segment-wise metric, we use the F1@50 as proposed by Lea et al. \cite{Lea2017}. to compute the segmental metric, a predicted action segment is first classified as a true positive (TP) or false positive (FP) by comparing its intersection over union (IoU) with respect to the corresponding expert annotation. If the IoU crosses a predetermined overlap threshold it is classified as a true positive segment (TP), if it does not, as a false positive segment (FP). The number of false-negative segments (FN) in a trial is calculated by subtracting the number of correctly predicted segments from the number of segments that the experts had demarcated. From the classified segments, the segmental F1-score for each action can be computed as:
\begin{equation}
    F1@\tau = \frac{TP}{TP+ \frac{1}{2}(FP+FN)},
\end{equation}
where $\tau$ denotes the IoU overlap. \\
For the sample-wise metric, we report the accuracy. The term sample-wise denotes that the metric is computed for each sample or timestep. Unlike the segment-wise metric, the sample-wise accuracy does not heavily penalize over-segmentation errors. Reporting both the sample-wise accuracy and the segment-wise F1@50 thus allows assessment of over-segmentation problems. The sample-wise accuracy is simply computed as the number of correctly classified samples divided by the total number of samples. \\
All use cases were evaluated by assessing the generalization of the models to previously unseen subjects. For the three larger action segmentation datasets, we used a fixed test/train partition. For the two smaller gait analysis datasets, we used a leave one subject out (LOSO) cross-validation approach. \\
The three high sample rate marker-based MoCap datasets were resampled to 50 Hz. No additional pre-processing was performed. A summary is provided in Table \ref{table_data}.

\subsection{Statistics}
We aim to determine if the differences in predictive performance between the five architectures is statistically significant. Several statistical methods have been proposed to compare machine learning algorithms \cite{Demsar2006-xi, Garcia2010-ed}. Demšar and Garcia et al. recommend the non-parametric Friedman test \cite{Friedman1937-gn}, with the corresponding post-hoc tests, for the comparison of more than two classifiers over multiple datasets or trials. We used Friedman's test to evaluate the null hypothesis that there is no difference in the classification performance of the five architectures on a particular dataset. The post-hoc tests were used to evaluate the null hypothesis that there is no difference in the classification performance between the proposed MS-GCN model and the four baselines on a particular dataset. The post-hoc hypotheses were corrected for multiple comparisons, as defined in Li \cite{Li2018-qw}. \\
All statistical analyses were performed using the scmamp package, version 0.2.55 \cite{scmamp}, within The R programming language, version 4.0.3 \cite{R_stat}. The scmamp package implemented Friedman's test according to the version by Demšar \cite{Demsar2006-xi} and the post-hoc tests according to the version by Garcia et al. \cite{Garcia2010-ed}. The significance level of all tests was set at the 95\% level ( p$\leq$0.05).

\section{Results}
\begin{table}[!t]
\caption{Action segmentation results.}
\centering
\begin{tabular}{lcc}
\textbf{PKU-MMD} & \textbf{F1@50} & \multicolumn{1}{l}{\textbf{Acc}} \\ \hline
Bi-LSTM             &  22.7               &   59.6                               \\
TCN                 &  13.8             &     61.9                             \\
ST-GCN              &   15.5             &   64.9                               \\
MS-TCN              &   46.3             &    65.5                              \\
MS-GCN              &   \textbf{51.6}    &    \textbf{68.5}                             \\ \\
\textbf{HuGaDB} & \textbf{F1@50} & \multicolumn{1}{l}{\textbf{Acc}} \\ \hline
Bi-LSTM         &    81.5            &    86.1                              \\
TCN             &   56.8             &     88.3                             \\
ST-GCN          &    67.7            &     88.7                             \\
MS-TCN          &    89.9            &      86.8                            \\
MS-GCN          &    \textbf{93.0}    &    \textbf{90.4}                             \\ \\
\textbf{LARa} & \textbf{F1@50} & \multicolumn{1}{l}{\textbf{Acc}}  \\ \hline
Bi-LSTM       &     32.3           &     63.9                             \\
TCN           &     20.0          &      61.5                            \\
ST-GCN        &    25.8            &     \textbf{67.9}                             \\
MS-TCN        &   39.6            &     65.8                           \\
MS-GCN        &   \textbf{43.6}             &    65.6                            \\ \\
\textbf{FOG-GAIT} & \textbf{F1@50} & \multicolumn{1}{l}{\textbf{Acc}} \\ \hline
Bi-LSTM       &     92.1           &    \textbf{90.6}                              \\
TCN           &     89.9           &    89.8                              \\
ST-GCN        &     90.8           &    89.4                              \\
MS-TCN        &     92.5           &    86.7                              \\
MS-GCN        &     \textbf{95.0}           &    90.1                             \\ \\
\textbf{TUG} & \textbf{F1@50} & \multicolumn{1}{l}{\textbf{Acc}} \\ \hline
Bi-LSTM      &    97.1            &    93.2               \\
TCN          &    84.4            &    92.7                   \\
ST-GCN       &    93.8            &    93.2              \\
MS-TCN       &    96.5            &    92.7            \\
MS-GCN       &    \textbf{97.9}  &    \textbf{93.6}        \\
\end{tabular}
\label{table_results}
\begin{tablenotes}
\item
Skeleton-based action segmentation results on PKU-MMD v2, HuGaDB, LARa, FOG-GAIT, and TUG. All results are quantified in terms of segment-wise F1@50 and sample-wise accuracy (Acc).
\end{tablenotes}
\end{table}

\begin{figure}
\centering
\includegraphics[width=3in]{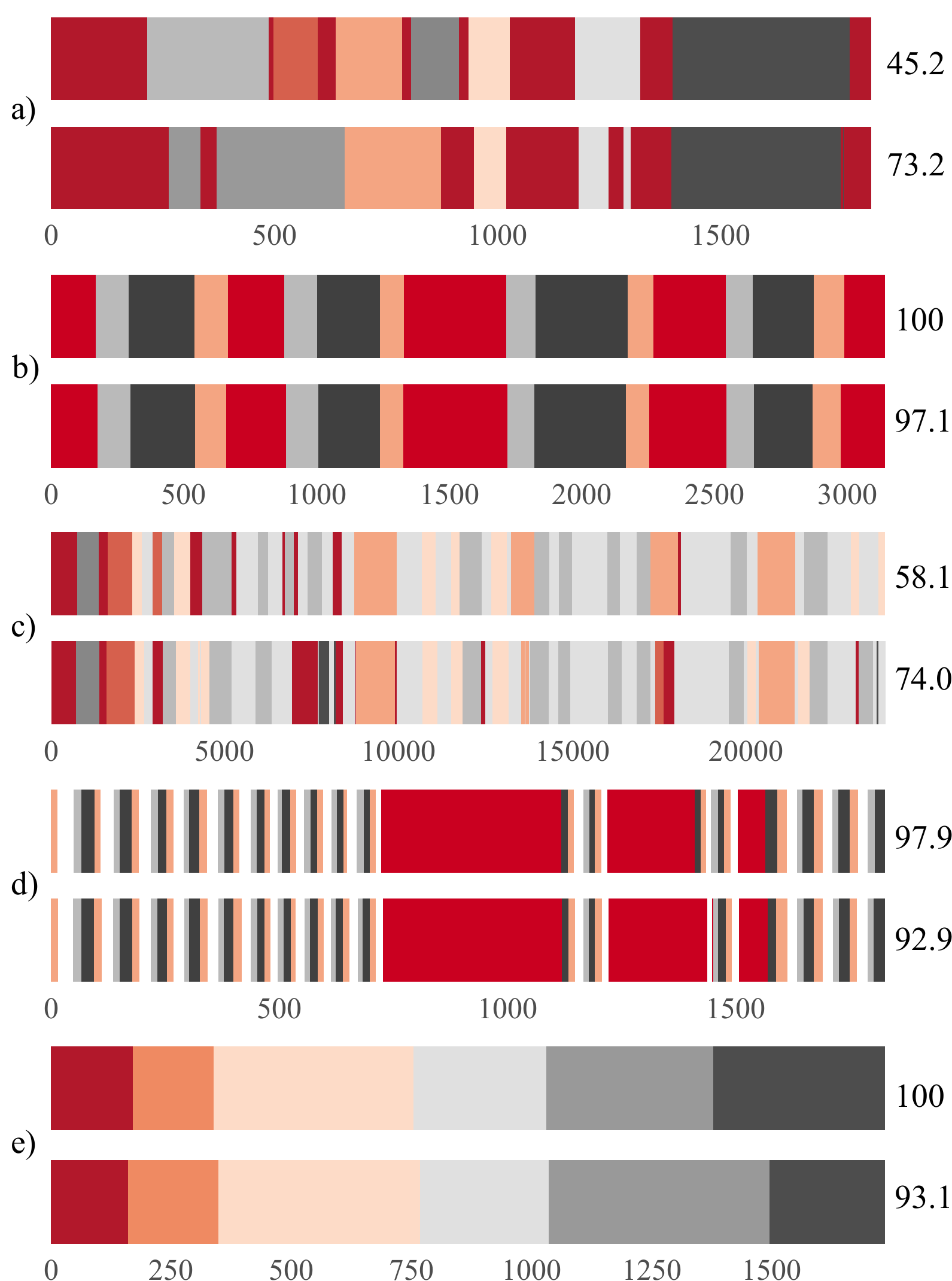}
\caption{Visual overview of action segmentation results for one sequence of each use-case. From top to bottom: (a) PKU-MMD v2, (b) HuGaDB, (c) LARa, (d) FOG-GAIT, and (e) TUG. For each use-case, the first sequence represents the GT and the second the segmentation by MS-GCN. For the visualized sequences, the segment-wise F1@50 score is given after the GT sequence and the sample-wise accuracy after the predicted sequence.}
\label{fig_segm}
\end{figure}
\subsection{Comparison with the four baselines}
\begin{figure*}[!t]
\centering
\includegraphics[width=7in]{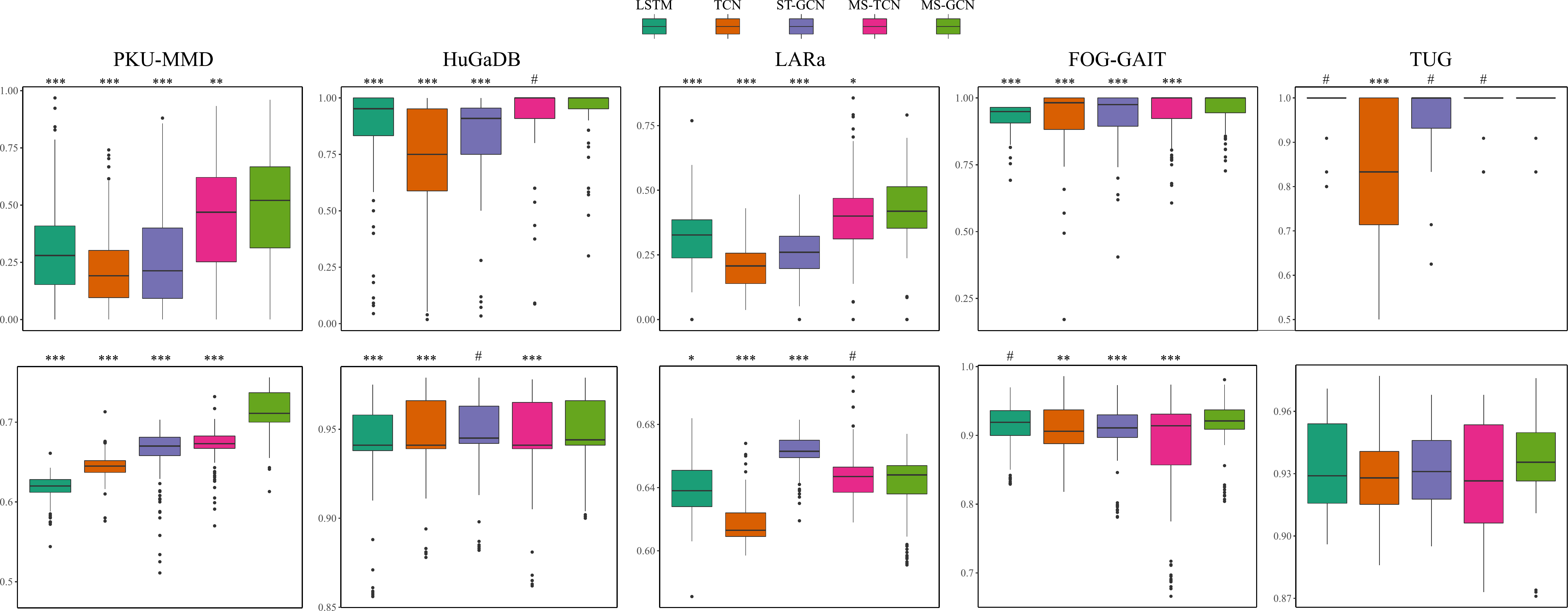}
\caption{Boxplots to visualize the spread in the segment-wise F1@50 (top row) and the sample-wise accuracy (bottom row) across trials per dataset. Significance levels were visualized as: p$\leq$0.01 (***), p$\leq$0.05 (**), p$\leq$0.1 (*), and no significance (\#). The Friedman test was significant at the p$\leq$0.01 (***) level for all but the TUG dataset, for which the significance of the F1@50 was found to be p$\leq$0.05 (**) and not significant for accuracy. The significance level of the post-hoc tests with respect to the MS-GCN model (corrected for multiple-comparisons) are visualized above their respective boxplot. No post-hoc tests were performed for the sample-wise accuracy of the TUG dataset since the Friedman test was not significant at the 95\% level (p$\leq$0.05 (**)).}
\label{fig_boxplots}
\end{figure*}

Results on all five datasets in terms of the segment-wise F1@50 and sample-wise accuracy (Acc) are shown in Table \ref{table_results}. Figure \ref{fig_segm} gives a visual overview of the segmentation results for MS-GCN. Sample-wise accuracy and segment-wise F1@50 for each sequence are included for comparison. The results of the statistical hypotheses tests and the spread across evaluation trials are visualized in Figure \ref{fig_boxplots}. All methods were evaluated in acausal mode.\\
The results suggest that MS-GCN outperforms the four baseline approaches across all tasks on most metrics. Figure \ref{fig_segm} indicates that MS-GCN enables a near perfect action segmentation on HuGaDB, FOG-GAIT, and TUG. The Friedman test was statistically significant at the 95\% level for both metrics on all but the TUG dataset, for which the accuracy was found to not be significant. We thus reject the null hypothesis that there is no difference in the classification performance among the five models on all but the TUG dataset. The post-hoc tests between MS-GCN and the second best model were statistically significant at the 95\% level level for the F1@50 metric on PKU-MMD and FOG-GAIT, and for the accuracy on PKU-MMD. For these tasks, we thus reject the null hypotheses that there is no difference in the classification performance between MS-GCN and the second best model. 

\subsection{Effect of the refinement stages}
Notice that the single-stage models (ST-GCN and TCN) and the multi-stage models (MS-GCN and MS-TCN) achieve similar sample-wise accuracy but very different F1@50 scores. The statistical tests confirm these observations, as the difference in F1@50 between ST-GCN and MS-GCN was statistically significant at the 95\% level across all datasets except for the TUG dataset, while the difference in sample-wise accuracy varied with MS-GCN performing significantly better (PKU-MMD and FOG-GAIT), no significant effect (HuGaDB and TUG), and ST-GCN performing significantly better (LARa). These results indicate that the addition of the refinement stages (i.e., multi-stage models) significantly reduces the number of segmentation errors. 

\subsection{Effect of the graph convolutions}
Notice that the graph convolutional models (ST-GCN and MS-GCN) outperform the regular convolutional models (TCN and MS-TCN) on all tasks. This effect was found to have a higher impact on the sample-wise accuracy than on the number of segmentation errors (F1@50). The statistical tests confirm these observations, as the difference between MS-GCN and MS-TCN was statistically significant at the 95\% level on two datasets (PKU-MMD and FOG-GAIT) for the F1@50 and on three datasets (PKU-MMD, HuGaDB, and FOG-GAIT) for the sample-wise accuracy. These results confirm that it is beneficial to explicitly model the spatial hierarchy among the joints or limbs in skeleton-based action segmentation tasks.

\subsection{Ablative experiments}
\subsubsection{Effect of the dilated convolutions}
Ablative experiments for MS-GCN were carried out to assess the effect of the introduced dilated temporal convolutions in the prediction generation stage. According to the results in Table \ref{table_dil}, it is evident that the introduced dilation within the ST-GCN layers of the prediction generation stage has a positive effect on both metrics across all datasets. The drop in performance with regular convolutions is due to the fact that without dilation the initial predictions are generated based on limited temporal context.

\begin{table}[!h]
\caption{MS-GCN: Effect of the dilated convolutions in the prediction generation stage.}
\centering
\begin{tabular}{lcc|cc}
 & \multicolumn{2}{c|}{Regular convolutions}  & \multicolumn{2}{c}{Dilated convolutions} \\ \hline
\textbf{Dataset} & \textbf{F1@50} & \textbf{Acc} & \textbf{F1@50} & \textbf{Acc} \\ \hline
PKU-MMD      &   44.8             &  68.4   & \textbf{51.6} & \textbf{68.5}                             \\
HuGaDB          &   75.5             &  83.8    & \textbf{93.0} & \textbf{90.4}                               \\
LARa              & 37.5 & 57.0 & \textbf{43.6} & \textbf{65.6} \\
FOG-GAIT & 88.1 & 85.7 & \textbf{95.0} & \textbf{90.1} \\
TUG & 85.8 & 90.0 & \textbf{97.9} & \textbf{93.6} \\\hline
\end{tabular}
\label{table_dil}
\end{table}

\subsubsection{Causal versus acausal convolutions}
We perform causal versus acausal experiments for MS-GCN. According to the results in Table \ref{table_causal}, MS-GCN with acausal temporal convolutions performs much better than the causal variant. The effect is larger on the segment-wise metric than the sample-wise metric. This verifies that future context is important for determining plausible action durations and accurate boundaries between action segments.  

\begin{table}[!h]
\caption{MS-GCN: Causal versus acausal temporal convolutions.}
\centering
\begin{tabular}{lcc|cc}
 & \multicolumn{2}{c|}{Causal convolutions}  & \multicolumn{2}{c}{Acausal convolutions} \\ \hline
\textbf{Dataset} & \textbf{F1@50} & \textbf{Acc} & \textbf{F1@50} & \textbf{Acc} \\ \hline
PKU-MMD      & 24.8              &  58.2   & \textbf{51.6} & \textbf{68.5}                             \\
HuGaDB          &  65.6             & 85.7    & \textbf{93.0} & \textbf{90.4}                               \\
LARa              & 18.5  &  57.0 & \textbf{43.6} & \textbf{65.6} \\
FOG-GAIT & 85.7 & 89.0 & \textbf{95.0} & \textbf{90.1} \\
TUG & 88.3 & 91.5 & \textbf{97.9} & \textbf{93.6} \\\hline
\end{tabular}
\label{table_causal}
\end{table}

\section{Conclusion}
This paper evaluated a neural network architecture for skeleton-based action segmentation, termed multi-stage spatial-temporal graph convolutional network (MS-GCN), that we initially developed for freezing of gait assessment in Parkinson's disease \cite{Filtjens2021-hu}. The developed architecture amalgamates three architectural elaborations based on the current best practices in convolutional neural network design. We presented five challenging use-cases of skeleton-based action segmentation in human action understanding and clinical gait analysis. The results indicated that our framework statistically outperformed four strong baselines on four of the five datasets. For the fifth dataset, i.e. the segmentation of TUG sub-activities, the task was found to be too simple, resulting in minimal to no statistical effect in the predictive performance of the models. The experimental evaluation demonstrated the benefit of the three architectural elaborations for detecting accurate action sequences with precise temporal boundaries. In conclusion, we believe that the MS-GCN framework is a formidable baseline for skeleton-based action segmentation tasks.  

\bibliographystyle{IEEEtran} 
\bibliography{bib}

\begin{thebibliography}{10}
\providecommand{\url}[1]{#1}
\csname url@samestyle\endcsname
\providecommand{\newblock}{\relax}
\providecommand{\bibinfo}[2]{#2}
\providecommand{\BIBentrySTDinterwordspacing}{\spaceskip=0pt\relax}
\providecommand{\BIBentryALTinterwordstretchfactor}{4}
\providecommand{\BIBentryALTinterwordspacing}{\spaceskip=\fontdimen2\font plus
\BIBentryALTinterwordstretchfactor\fontdimen3\font minus
  \fontdimen4\font\relax}
\providecommand{\BIBforeignlanguage}[2]{{%
\expandafter\ifx\csname l@#1\endcsname\relax
\typeout{** WARNING: IEEEtran.bst: No hyphenation pattern has been}%
\typeout{** loaded for the language `#1'. Using the pattern for}%
\typeout{** the default language instead.}%
\else
\language=\csname l@#1\endcsname
\fi
#2}}
\providecommand{\BIBdecl}{\relax}
\BIBdecl

\bibitem{Al-Amri2018-yd}
M.~Al-Amri, K.~Nicholas, K.~Button, V.~Sparkes, L.~Sheeran, and J.~L. Davies,
  ``\BIBforeignlanguage{en}{Inertial measurement units for clinical movement
  analysis: Reliability and concurrent validity},''
  \emph{\BIBforeignlanguage{en}{Sensors}}, vol.~18, no.~3, Feb. 2018.

\bibitem{Mahmood2019-au}
N.~Mahmood, N.~Ghorbani, N.~F. Troje, G.~Pons-Moll, and M.~Black,
  ``\BIBforeignlanguage{en}{{AMASS}: Archive of motion capture as surface
  shapes},'' in \emph{\BIBforeignlanguage{en}{2019 {IEEE/CVF} International
  Conference on Computer Vision ({ICCV})}}.\hskip 1em plus 0.5em minus
  0.4em\relax IEEE, Oct. 2019.

\bibitem{Shotton2013-zx}
J.~Shotton, R.~Girshick, A.~Fitzgibbon, T.~Sharp, M.~Cook, M.~Finocchio,
  R.~Moore, P.~Kohli, A.~Criminisi, A.~Kipman, and A.~Blake,
  ``\BIBforeignlanguage{en}{Efficient human pose estimation from single depth
  images},'' \emph{\BIBforeignlanguage{en}{IEEE Trans. Pattern Anal. Mach.
  Intell.}}, vol.~35, no.~12, pp. 2821--2840, Dec. 2013.

\bibitem{Cao2018-yi}
Z.~{Cao}, G.~{Hidalgo}, T.~{Simon}, S.~E. {Wei}, and Y.~{Sheikh}, ``Openpose:
  Realtime multi-person 2d pose estimation using part affinity fields,''
  \emph{IEEE Transactions on Pattern Analysis and Machine Intelligence},
  vol.~43, no.~1, pp. 172--186, 2021.

\bibitem{Fernando2015-hb}
B.~Fernando, E.~Gavves, M.~Jos{\'e}~Oramas, A.~Ghodrati, and T.~Tuytelaars,
  ``Modeling video evolution for action recognition,'' in \emph{2015 {IEEE}
  Conference on Computer Vision and Pattern Recognition ({CVPR})}, Jun. 2015,
  pp. 5378--5387.

\bibitem{Shahroudy2016-sz}
A.~Shahroudy, J.~Liu, T.-T. Ng, and G.~Wang, ``{NTU} {RGB+D}: A large scale
  dataset for {3D} human activity analysis,'' pp. 1010--1019, Apr. 2016.

\bibitem{Yan2018-jp}
S.~Yan, Y.~Xiong, and D.~Lin, ``Spatial temporal graph convolutional networks
  for skeleton-based action recognition,'' in \emph{AAAI}, 2018.

\bibitem{Defferrard2016-cb}
M.~Defferrard, X.~Bresson, and P.~Vandergheynst, ``Convolutional neural
  networks on graphs with fast localized spectral filtering,'' in
  \emph{Proceedings of the 30th International Conference on Neural Information
  Processing Systems}, ser. NIPS'16.\hskip 1em plus 0.5em minus 0.4em\relax Red
  Hook, NY, USA: Curran Associates Inc., Dec. 2016, pp. 3844--3852.

\bibitem{Kipf2017}
T.~N. Kipf and M.~Welling, ``{Semi-Supervised Classification with Graph
  Convolutional Networks},'' \emph{International conference on learning
  representation}, p.~14, 2017.

\bibitem{Singh2016-iy}
B.~Singh, T.~K. Marks, M.~Jones, O.~Tuzel, and M.~Shao, ``A multi-stream
  bi-directional recurrent neural network for {Fine-Grained} action
  detection,'' in \emph{2016 {IEEE} Conference on Computer Vision and Pattern
  Recognition ({CVPR})}, Jun. 2016, pp. 1961--1970.

\bibitem{Sun2015-ny}
L.~Sun, K.~Jia, D.-Y. Yeung, and B.~E. Shi, ``Human action recognition using
  factorized {Spatio-Temporal} convolutional networks,'' in \emph{2015 {IEEE}
  International Conference on Computer Vision ({ICCV})}, Dec. 2015, pp.
  4597--4605.

\bibitem{Yao2018-iz}
R.~Yao, G.~Lin, Q.~Shi, and D.~C. Ranasinghe, ``Efficient dense labelling of
  human activity sequences from wearables using fully convolutional networks,''
  \emph{Pattern Recognit.}, vol.~78, pp. 252--266, Jun. 2018.

\bibitem{Lea2017}
C.~Lea, M.~D. Flynn, R.~Vidal, A.~Reiter, and G.~D. Hager, ``{Temporal
  convolutional networks for action segmentation and detection},''
  \emph{Proceedings - 30th IEEE Conference on Computer Vision and Pattern
  Recognition, CVPR 2017}, vol. 2017-January, pp. 1003--1012, 2017.

\bibitem{Yu2015-qu}
F.~Yu and V.~Koltun, ``{Multi-Scale} context aggregation by dilated
  convolutions,'' \emph{pre-print}, Nov. 2015.

\bibitem{Farha2019-yw}
Y.~A. {Farha} and J.~{Gall}, ``Ms-tcn: Multi-stage temporal convolutional
  network for action segmentation,'' in \emph{2019 IEEE/CVF Conference on
  Computer Vision and Pattern Recognition (CVPR)}, 2019, pp. 3570--3579.

\bibitem{Filtjens2021-hu}
B.~Filtjens, P.~Ginis, A.~Nieuwboer, P.~Slaets, and B.~Vanrumste,
  ``\BIBforeignlanguage{en}{Automated freezing of gait assessment with
  marker-based motion capture and multi-stage spatial-temporal graph
  convolutional neural networks},'' \emph{\BIBforeignlanguage{en}{J. Neuroeng.
  Rehabil.}}, vol.~19, no.~1, p.~48, May 2022.

\bibitem{Goodfellow2016-tq}
I.~Goodfellow, Y.~Bengio, and A.~Courville, \emph{Deep Learning}.\hskip 1em
  plus 0.5em minus 0.4em\relax The MIT Press, 2016.

\bibitem{Lecun1998-an}
Y.~Lecun, L.~Bottou, Y.~Bengio, and P.~Haffner, ``Gradient-based learning
  applied to document recognition,'' \emph{Proc. IEEE}, vol.~86, no.~11, pp.
  2278--2324, Nov. 1998.

\bibitem{Bai2018-uk}
S.~Bai, J.~Zico~Kolter, and V.~Koltun, ``An empirical evaluation of generic
  convolutional and recurrent networks for sequence modeling,'' Mar. 2018.

\bibitem{Graves2005-gv}
A.~Graves and J.~Schmidhuber, ``\BIBforeignlanguage{en}{Framewise phoneme
  classification with bidirectional {LSTM} and other neural network
  architectures},'' \emph{\BIBforeignlanguage{en}{Neural Netw.}}, vol.~18, no.
  5-6, pp. 602--610, Jun. 2005.

\bibitem{Filtjens2020-hl}
B.~Filtjens, A.~Nieuwboer, N.~D'cruz, J.~Spildooren, P.~Slaets, and
  B.~Vanrumste, ``A data-driven approach for detecting gait events during
  turning in people with parkinson's disease and freezing of gait,'' \emph{Gait
  Posture}, vol.~80, pp. 130--136, Jul. 2020.

\bibitem{Matsushita2021-kx}
Y.~Matsushita, D.~T. Tran, H.~Yamazoe, and J.-H. Lee, ``Recent use of deep
  learning techniques in clinical applications based on gait: a survey,''
  \emph{Journal of Computational Design and Engineering}, vol.~8, no.~6, pp.
  1499--1532, Oct. 2021.

\bibitem{Cheema2018-sh}
N.~Cheema, S.~Hosseini, J.~Sprenger, E.~Herrmann, H.~Du, K.~Fischer, and
  P.~Slusallek, ``Dilated temporal {Fully-Convolutional} network for semantic
  segmentation of motion capture data,'' Jun. 2018.

\bibitem{Ioffe2015-ta}
S.~Ioffe and C.~Szegedy, ``Batch normalization: Accelerating deep network
  training by reducing internal covariate shift,'' in \emph{Proceedings of the
  32nd International Conference on International Conference on Machine Learning
  (ICML)}, 2015.

\bibitem{Schuster1997-nd}
M.~Schuster and K.~K. Paliwal, ``Bidirectional recurrent neural networks,''
  \emph{Trans. Sig. Proc.}, vol.~45, no.~11, pp. 2673--2681, Nov. 1997.

\bibitem{Kidzinski2019-ou}
{\L}.~Kidzi{\'n}ski, S.~Delp, and M.~Schwartz,
  ``\BIBforeignlanguage{en}{Automatic real-time gait event detection in
  children using deep neural networks},'' \emph{\BIBforeignlanguage{en}{PLoS
  One}}, vol.~14, no.~1, p. e0211466, Jan. 2019.

\bibitem{Kingma2014-va}
D.~P. Kingma and J.~Ba, ``Adam: A method for stochastic optimization,''
  \emph{pre-print}, Dec. 2014.

\bibitem{Liu2017-jx}
C.~Liu, Y.~Hu, Y.~Li, S.~Song, and J.~Liu, ``{PKU-MMD}: A large scale benchmark
  for {Skeleton-Based} human action understanding,'' in \emph{Proceedings of
  the Workshop on Visual Analysis in Smart and Connected Communities}, ser.
  VSCC '17.\hskip 1em plus 0.5em minus 0.4em\relax New York, NY, USA:
  Association for Computing Machinery, Oct. 2017, pp. 1--8.

\bibitem{Chereshnev2017HuGaDBHG}
R.~Chereshnev and A.~Kert{\'e}sz-Farkas, ``Hugadb: Human gait database for
  activity recognition from wearable inertial sensor networks,'' in
  \emph{AIST}, 2017.

\bibitem{Niemann2020-ut}
F.~Niemann, C.~Reining, F.~Moya~Rueda, N.~R. Nair, J.~A. Steffens, G.~A. Fink,
  and M.~Ten~Hompel, ``\BIBforeignlanguage{en}{{LARa}: Creating a dataset for
  human activity recognition in logistics using semantic attributes},''
  \emph{\BIBforeignlanguage{en}{Sensors}}, vol.~20, no.~15, Jul. 2020.

\bibitem{Spildooren2010-pj}
J.~Spildooren, S.~Vercruysse, K.~Desloovere, W.~Vandenberghe, E.~Kerckhofs, and
  A.~Nieuwboer, ``\BIBforeignlanguage{en}{Freezing of gait in parkinson's
  disease: the impact of dual-tasking and turning},''
  \emph{\BIBforeignlanguage{en}{Mov. Disord.}}, vol.~25, no.~15, pp.
  2563--2570, Nov. 2010.

\bibitem{Nieuwboer2001-cr}
A.~Nieuwboer, R.~Dom, W.~De~Weerdt, K.~Desloovere, S.~Fieuws, and
  E.~Broens-Kaucsik, ``\BIBforeignlanguage{en}{Abnormalities of the
  spatiotemporal characteristics of gait at the onset of freezing in
  parkinson's disease},'' \emph{\BIBforeignlanguage{en}{Mov. Disord.}},
  vol.~16, no.~6, pp. 1066--1075, Nov. 2001.

\bibitem{Schaafsma2003-pz}
J.~D. Schaafsma, Y.~Balash, T.~Gurevich, A.~L. Bartels, J.~M. Hausdorff, and
  N.~Giladi, ``\BIBforeignlanguage{en}{Characterization of freezing of gait
  subtypes and the response of each to levodopa in parkinson's disease},''
  \emph{\BIBforeignlanguage{en}{Eur. J. Neurol.}}, vol.~10, no.~4, pp.
  391--398, Jul. 2003.

\bibitem{Gilat2019-xm}
M.~Gilat, ``\BIBforeignlanguage{en}{How to annotate freezing of gait from
  video: A standardized method using {Open-Source} software},''
  \emph{\BIBforeignlanguage{en}{J. Parkinsons. Dis.}}, vol.~9, no.~4, pp.
  821--824, 2019.

\bibitem{Bowen2001-uc}
A.~Bowen, R.~Wenman, J.~Mickelborough, J.~Foster, E.~Hill, and R.~Tallis,
  ``\BIBforeignlanguage{en}{Dual-task effects of talking while walking on
  velocity and balance following a stroke},'' \emph{\BIBforeignlanguage{en}{Age
  Ageing}}, vol.~30, no.~4, pp. 319--323, Jul. 2001.

\bibitem{Davis1991-tr}
R.~B. Davis, S.~{\~O}unpuu, D.~Tyburski, and J.~R. Gage, ``A gait analysis data
  collection and reduction technique,'' \emph{Hum. Mov. Sci.}, vol.~10, no.~5,
  pp. 575--587, Oct. 1991.

\bibitem{Podsiadlo1991-gy}
D.~Podsiadlo and S.~Richardson, ``\BIBforeignlanguage{en}{The timed ``up \&
  go'': a test of basic functional mobility for frail elderly persons},''
  \emph{\BIBforeignlanguage{en}{J. Am. Geriatr. Soc.}}, vol.~39, no.~2, pp.
  142--148, Feb. 1991.

\bibitem{Li2018-qw}
T.~Li, J.~Chen, C.~Hu, Y.~Ma, Z.~Wu, W.~Wan, Y.~Huang, F.~Jia, C.~Gong, S.~Wan,
  and L.~Li, ``Automatic timed {Up-and-Go} {Sub-Task} segmentation for
  parkinson's disease patients using {Video-Based} activity classification,''
  \emph{IEEE Trans. Neural Syst. Rehabil. Eng.}, vol.~26, no.~11, pp.
  2189--2199, Nov. 2018.

\bibitem{Liang2020-hq}
P.~Liang, W.~H. Kwong, A.~Sidarta, C.~K. Yap, W.~K. Tan, L.~S. Lim, P.~Y. Chan,
  C.~W.~K. Kuah, S.~K. Wee, K.~Chua, C.~Quek, and W.~T. Ang,
  ``\BIBforeignlanguage{en}{An asian-centric human movement database capturing
  activities of daily living},'' \emph{\BIBforeignlanguage{en}{Sci Data}},
  vol.~7, no.~1, p. 290, Sep. 2020.

\bibitem{Cappozzo1995-ra}
A.~Cappozzo, F.~Catani, U.~D. Croce, and A.~Leardini,
  ``\BIBforeignlanguage{en}{Position and orientation in space of bones during
  movement: anatomical frame definition and determination},''
  \emph{\BIBforeignlanguage{en}{Clin. Biomech.}}, vol.~10, no.~4, pp. 171--178,
  Jun. 1995.

\bibitem{Demsar2006-xi}
J.~Dem{\v s}ar, ``Statistical comparisons of classifiers over multiple data
  sets,'' \emph{J. Mach. Learn. Res.}, vol.~7, no.~1, pp. 1--30, 2006.

\bibitem{Garcia2010-ed}
S.~Garc{\'\i}a, A.~Fern{\'a}ndez, J.~Luengo, and F.~Herrera, ``Advanced
  nonparametric tests for multiple comparisons in the design of experiments in
  computational intelligence and data mining: Experimental analysis of power,''
  \emph{Inf. Sci.}, vol. 180, no.~10, pp. 2044--2064, May 2010.

\bibitem{Friedman1937-gn}
M.~Friedman, ``The use of ranks to avoid the assumption of normality implicit
  in the analysis of variance,'' \emph{J. Am. Stat. Assoc.}, vol.~32, no. 200,
  pp. 675--701, Dec. 1937.

\bibitem{scmamp}
C.~Borja and S.~Guzman, ``scmamp: Statistical comparison of multiple algorithms
  in multiple problems,'' \emph{The R Journal}, vol. Accepted for publication,
  2015.

\bibitem{R_stat}
\BIBentryALTinterwordspacing
{R Core Team}, \emph{R: A Language and Environment for Statistical Computing},
  R Foundation for Statistical Computing, Vienna, Austria, 2013. [Online].
  Available: \url{http://www.R-project.org/}
\BIBentrySTDinterwordspacing

\end{thebibliography}

\begin{IEEEbiographynophoto}{Benjamin Filtjens}
received a MSc in Mechanical Engineering Technology from Hasselt University in 2017. He is currently a Ph.D. student working towards automated at-home freezing of gait assessment, at KU Leuven. He is part of the eMedia research lab at the Department of Electrical Engineering (ESAT) and the intelligent mobile platform research group at the Department of Mechanical Engineering, both from KU Leuven campus Group T. At Group T, he teaches mathematical modelling, advanced automation engineering, and deep learning in health technologies. His research interests are deep learning, explainable artificial intelligence, and mobile robots in general, and ICT applications for automatic and objective gait and freezing of gait assessment in particular.
\end{IEEEbiographynophoto}

\begin{IEEEbiographynophoto}{Bart Vanrumste}
received a MSc in Electrical Engineering and MSc in
Biomedical Engineering both from Ghent University in 1994 and 1998,
respectively. In 2001 he received a Ph.D. in Engineering from the same
institute. He worked as a post-doctoral fellow from 2001 until 2003 at the
Electrical and Computer Engineering Department of the University Of
Canterbury, New Zealand. From 2003 until 2005 he was post-doctoral
fellow at the Department of Electrical Engineering (ESAT) in the STADIUS division at KU Leuven. In 2005 he was appointed faculty member
initially at University of Applied Sciences Thomas More and since 2013
in the Faculty of Engineering Technology of KU Leuven. He is member of
the eMedia research lab and member of the ESAT-STADIUS division. His current research activities focus on multimodal sensor integration. He is senior member of IEEE Engineering in Medicine and Biology Society.
\end{IEEEbiographynophoto}

\begin{IEEEbiographynophoto}{Peter Slaets}
received a MSc in electrotechnical-mechanical engineering, specialization in datamining and automation, in 2002 from KU Leuven, Leuven, Belgium. In 2005, he became a lecturer at the Katholieke Hogeschool Limburg (KHLIM), Diepenbeek, Belgium, and the Katholieke Hogeschool Kempen (KHK), Geel, Belgium, where he teaches courses in digital electronics, control, and automation. In 2008 He received a Ph.D. in applied sciences from KU Leuven with the title: 'Geometric 3D Model Building from Sensor Measurements Collected during Compliant Motion: Stochastic Filtering and Hardware Architectures'. He is currently an associate professor in the intelligent mobile platform research group at the Department of Mechanical Engineering, KU Leuven. His research focuses on modeling, Bayesian estimation techniques, and mobile platforms in general, with applications in autonomous inland shipping and health monitoring in particular.
\end{IEEEbiographynophoto}

\end{document}